\pdfoutput=1
\documentclass[11pt,a4paper]{article}
\usepackage[hyperref]{emnlp2021}
\usepackage{times}
\usepackage{latexsym}
\usepackage{color}
\usepackage{multirow}
\usepackage{forest}
\usepackage{tikz}
\usepackage{algorithm}
\usepackage{algpseudocode}
\usepackage{amsmath}
\usepackage{graphics}
\usepackage{subfigure}
\usepackage{mwe}
\usepackage{amssymb}
\usepackage{bm}
\usepackage{amsmath}
\usepackage{makecell}
\usepackage{tabularx}
\usepackage{url}
\usepackage{pgfplots}
\usepackage{CJKutf8}
\pgfplotsset{width=10cm,compat=1.9}
\usepackage{booktabs}
\usepackage{dsfont}
\usepackage{arydshln}
\usepackage{enumitem}

\usepackage{microtype}

\definecolor{electricviolet}{rgb}{0.56, 0.0, 1.0}

\usetikzlibrary{arrows.meta,calc,decorations.markings,math,arrows.meta,patterns}
\usetikzlibrary{decorations.pathreplacing}
\usetikzlibrary{matrix}

\title{On the Complementarity between Pre-Training and Back-Translation \\ for Neural Machine Translation}

\author{Xuebo Liu$^{1}$\thanks{~~Work was done when Xuebo Liu and Liang Ding were interning at Tencent AI Lab.} , Longyue Wang$^{2}$, Derek F. Wong$^{1}$, Liang Ding$^{3}$, \\ \bf Lidia S. Chao$^{1}$, Shuming Shi$^{2}$ and Zhaopeng Tu$^{2}$\\
  $^{1}$NLP$^2$CT Lab, Department of Computer and Information Science, 
  University of Macau \\
  $^{2}$Tencent AI Lab~~~~$^{3}$The University of Sydney \\
  $^{1}${\tt nlp2ct.xuebo@gmail.com, \{derekfw,lidiasc\}@um.edu.com} \\
  $^{2}${\tt \{vinnylywang,shumingshi,zptu\}@tencent.com} \\
  $^{3}${\tt ldin3097@sydney.edu.au}}

\date{}

\begin{document}
\maketitle
\begin{abstract}
Pre-training (PT) and back-translation (BT) are two simple and powerful methods to utilize monolingual data for improving the model performance of neural machine translation (NMT).
This paper takes the first step to investigate the complementarity between PT and BT.
We introduce two probing tasks for PT and BT respectively and find that PT mainly contributes to the encoder module while BT brings more benefits to the decoder.
Experimental results show that PT and BT are nicely complementary to each other, establishing state-of-the-art performances on the WMT16 English-Romanian and English-Russian benchmarks.
Through extensive analyses on sentence originality and word frequency, we also demonstrate that combining Tagged BT with PT is more helpful to their complementarity, leading to better translation quality.
Source code is freely available at \url{https://github.com/SunbowLiu/PTvsBT}.
\end{abstract}

\section{Introduction}
Neural machine translation (\citealp[NMT;][]{bahdanau2014neural,DBLP:journals/corr/GehringAGYD17,DBLP:journals/corr/VaswaniSPUJGKP17}) models are data-hungry and their performances are highly dependent upon the quantity and quality of labeled data, which are expensive and scarce resources~\citep{Leong-etal-exploiting}.
This motivates the research line of exploiting unlabeled monolingual data for boosting the model performance of NMT.
Due to simplicity and effectiveness, pre-training (\citealp[PT;][]{Devlin:2018uk,song2019mass}) and back-translation (\citealp[BT;][]{sennrich2016improving}) are two widely-used techniques for NMT, by leveraging a large amount of monolingual data.

While empirically successful, the understandings of PT and BT are still limited at best. 
Several attempts have been made to better understand them at the data level, e.g. exploring different kinds of noises for the source data~\citep{edunov-etal-2018-understanding,bart2020}. 
However, there are few understandings at the model level that how PT and BT affect the internal module (e.g. encoder and decoder) of NMT models. 
As recent studies start to combine PT and BT for better model performance~\citep{lample2019cross,Liu:2020mbart,ding-etal-2021-improving}, there is a pressing need to broaden the understandings of them. 

To this end, we introduce two probing tasks to investigate the effects of PT and BT on the encoder and decoder modules, respectively. 
We find that PT mainly contributes to the encoder module while BT brings more benefits to the decoder module. 
This provides a good explanation for the performance improvement of simply combining PT and BT. 
Motivated by this finding, we explore a better combination method by leveraging {\it Tagged BT}~\citep{caswell2019tagged}. 
Experiments conducted on the WMT16 English-Romanian and English-Russian benchmarks show that PT can nicely co-work with BT, leading to state-of-the-art model performances. 
Extensive analyses show that the tagging mechanism is helpful for enhancing the complementarity between PT and BT by improving the translation of source-original sentences and low-frequency words. 

Our {\bf main contributions} are as follows:
\begin{itemize}
\item We design two probing tasks to investigate the impact of PT and BT on NMT models.

\item We empirically demonstrate the complementarity between PT and BT.

\item We show that Tagged BT further improves the complementarity between PT and BT.
\end{itemize}

\section{Preliminaries}

\subsection{Background}

\paragraph{Pre-Training for NMT}
Self-supervised PT~\citep{Devlin:2018uk,song2019mass}, which can acquire knowledge from unlabeled monolingual data, has shown its effectiveness in improving the model performance of NMT, especially for those language pairs with smaller parallel corpora~\citep{lample2019cross}.

The first research line treats pre-trained models as external knowledge to guidance NMT to learn better representations~\citep{yang2020towards,Zhu2020Incorporating} and predictions~\citep{chen2020distilling}.
These methods are effective but costly since the NMT architecture needed to be elaborately designed.
Another research line is directly taking the weights of pre-trained models to initialize NMT models, which is easy to use and advancing the state-of-the-art~\citep{rothe2020leveraging,bart2020}.
In this paper, we treat pre-trained mBART~\citep{Liu:2020mbart} as our testbed for parameter initialization, whose benefits have been sufficiently validated~\citep{tran2020cross,tang2020multilingual,liu-etal-2021-copying} by multiple translation directions.

In general, previous studies focus on designing novel architectures~\citep{song2019mass} and artificial noises for source sentences~\citep{lin2020pre,yang2020csp} but are still unclear why pre-training can boost the model performance of NMT, which is this paper aims to investigate.

\paragraph{Back-Translation for NMT}
BT is an alternative to leverage monolingual data for NMT~\citep{sennrich2016improving}.
It first trains a reversed NMT model for translating target-side monolingual data into synthetic parallel data, and then complements them with the original parallel data to train the desired NMT model.
To improve BT, previous works put attention to the importance of diversity and complexity in synthetic data, showing that adding symbols (e.g., noises and tags) to the back-translated source can help NMT distinguish the data from various sources and learn better representations~\citep{fadaee2018back,wang2019improving,edunov-etal-2018-understanding,caswell2019tagged,marie2020tagged}. 
The claims and understandings from these works are chiefly at the data-level rather than the model-level.

There also exists some works that combine PT and BT to further boost the model performance~\citep{conneau2019unsupervised,song2019mass,Liu:2020mbart}.
However, the relation between BT and PT has not been fully studied.
In this paper, we take the first step to understand BT and PT at the model-level and improve the complementarity between PT and BT.

\subsection{Experimental Setup}
\label{sec:setup}
\paragraph{Data}
We conducted experiments on the WMT16 English-Romanian (En-Ro) and English-Russia (En-Ru) translation tasks, which are widely-used benchmarks of data augmentation methods for NMT.
The training/validation/test sets of the En-Ro include 612K/2K/2K sentence pairs, while those of En-Ru include 2M/3K/3K pairs.
Towards better reproducibility, we directly used the BT data provided by~\citet{DBLP:journals/corr/SennrichHB16}\footnote{\url{http://data.statmt.org/rsennrich/wmt16_backtranslations}}, consisting of 2.3M synthetic data for the En-Ro and 2.0M data for the En-Ru.
All the data are tokenized and split into sub-words~\citep{DBLP:journals/corr/SennrichHB15} by the mBART tokenizer~\citep{Liu:2020mbart}.

\paragraph{Setting}
To make a fair comparison, all the model architectures and parameters are the same as the pre-trained {\tt mBART.cc25}.\footnote{\url{https://github.com/pytorch/fairseq/tree/master/examples/mbart}}
The NMT model augmented with PT directly uses the mBART weights for parameter initialization, while the other models randomly initialize their parameters.
The training follows~\citet{Liu:2020mbart} except that we tuned the learning rate within [3e-5,1e-3] and the dropout within [0.3,0.5] for the vanilla model and BT models.
We used the single model with the best validation perplexity for testing.
The length penalty is 1.0 and the beam size is 5.
We used sacreBLEU~\citep{post-2018-call} to calculate BLEU~\citep{papineni2002bleu} and TER~\citep{snover2006study} scores with the specific tokenization~\citep{Liu:2020mbart} for Romanian and the default tokenization for Russian.

\section{Understanding PT and BT}
\label{sec:understand}
In this section, we aim to better understand the similarities and differences between PT and BT on improving model performance.
We design two probing tasks to study the research question: {\em Which module of NMT do PT and BT respectively play a greater role in enhancing translation quality?}

\begin{table}[t]
\centering
\setlength{\tabcolsep}{5pt}
\begin{tabular}{cccrrr}
\toprule
\multirow{2}{*}{\bf Enc.}& \multirow{2}{*}{\bf Dec.}& \multicolumn{2}{c}{\bf PT} & \multicolumn{2}{c}{\bf BT} \\
\cmidrule(lr){3-4} \cmidrule(lr){5-6} 
 & &\bf BLEU &\bf $\Delta$  &\bf BLEU &\bf $\Delta$ \\
\midrule
N & N &33.7 &-    &33.7 &- \\ \hdashline 
N & Y &33.5 &-0.2    &\bf 37.8 &\bf +4.1 \\ 
Y & N &\bf 36.9 &\bf +3.2    &35.8 &+2.1 \\ \hdashline 
Y & Y &37.7 &+4.0    &38.3 &+4.6 \\   
\bottomrule
\end{tabular}
\caption{The probing tasks of PT and BT. NMT models are trained and evaluated on the WMT16 En-Ro benchmark. ``Y'' denotes the corresponding parameters are activated when augmented with PT or BT, while ``N'' denotes the inactive operation. PT and BT respectively contribute more to the NMT encoder and decoder.}
\label{tab:ablation}
\end{table}

\subsection{Effects of PT on NMT}
\label{sec:pteffect}
Given a pre-trained model, it is common to use its part or all parameters to initialize the downstream tasks. 
We design four NMT models, which differ from the NMT components (Encoder vs. Decoder) with parameter initialization manners (Random vs. Pre-trained).
As shown in Table~\ref{tab:ablation}, the variants are: 1) {\bf NN} is a vanilla NMT model, of which parameters are randomly initialized; 2) {\bf NY} means that parameters of NMT encoder are randomly initialized while those of decoder are initialized with pre-training; 3) {\bf YN} is contrary to {\bf NY}; and 4) {\bf YY} indicates that the whole NMT parameters are initialized with the pre-trained model. 
After that, the NMT models are fine-tuned on the parallel data with the same training strategy.

\paragraph{PT Mainly Contributes to Encoder}
As seen, the {\bf YY} initialization strategy significantly improves the vanilla NMT model by +4.0 BLEU scores, which reconfirms the effectiveness of PT for translation tasks~\cite{Liu:2020mbart}. By comparing {\bf NY} and {\bf YN}, we find that the pre-trained encoder can still help the NMT to achieve +3.2 BLEU improvements while the pre-trained decoder can only perform on par with the vanilla model (i.e. -0.2 BLEU). This demonstrates that PT mainly contributes to the encoder part of NMT model, and this claim is consistent with the conclusion with other pre-trained models. For instance, ~\citet{rothe2020leveraging} show that the NMT encoder initialization is superior to the decoder one when using pre-trained weights of BERT.
We hypothesize that the performance boost with PT mainly comes from the better ability of source-side understanding, which is significant to NMT such as on disambiguating word senses~\cite{tang2019encoders}.

\subsection{Effects of BT on NMT}
\label{sec:bteffect}
A vanilla NMT model is trained on the original bi-text and then fine-tuned on the mixture of the original and synthetic (i.e. back-translated) data.
We also design four NMT models, which differ from which parts of parameters are updated at the fine-tuning stage. 
As shown in Table~\ref{tab:ablation}, the variants are: 1) {\bf NN} is a vanilla NMT model only trained on the original data; 2) {\bf NY} indicates that parameters of the NMT encoder are fixed while those of decoder are updated during model fine-tuning; 3) {\bf YN} acts in an opposite way compared with {\bf NY}; 4) {\bf YY} means that the whole NMT parameters are updated at the fine-tuning stage. 

\paragraph{BT Mainly Contributes to Decoder}
BT has been sufficiently validated to improve the performance of NMT models ~\citep{edunov-etal-2018-understanding, edunov2020evaluation}. 
By exploiting additional target sentences, the NMT decoder can be enhanced to generate more fluent sentences in the target language. 
In contrast, the synthetic source sentences contain noises, which may be less useful for improving the ability of understanding. 
The results verify our hypothesis: BT mainly improves the decoder module of NMT. As seen, fine-tuning the whole NMT model (i.e. {\bf YY}) with BT data can gain the best performance (+4.6 BLEU than the vanilla model), which shows the effectiveness of BT method. 
Surprisingly, only fine-tuning the decoder (i.e. {\bf NY}) can perform close to {\bf YY} model (37.8 vs. 38.3 BLEU), which confirms our claims. Compared with {\bf NY}, the {\bf YN} model obtains relatively fewer improvements (+4.1 vs. +2.1 BLEU), showing that BT brings more benefits to the decoder than the encoder.

\section{Improving PT and BT}
The answer of the research question in Section~\ref{sec:understand} is: {\em PT and BT respectively contribute more to the NMT encoder and decoder, demonstrating that they are orthogonal and complementary to each other.}
This finding motivates us to better combine these two individual techniques together for further improving NMT models.

\subsection{Experiments}
As detailed in Section~\ref{sec:setup}, we conducted experiments on two commonly-used benchmarks En-Ro and En-Ru. Besides, we train the BT models from scratch instead of fine-tuning in Section~\ref{sec:bteffect}. As {\bf YY} models (in Table~\ref{tab:ablation}) always achieve best performances when augmented PT or BT, we update all parameters of NMT models in next experiments.

\begin{table}[t]
\centering
\setlength{\tabcolsep}{3pt}
\scalebox{0.92}{
\begin{tabular}{lrrrr}
\toprule
\multirow{2}{*}{\bf Model} & \multicolumn{2}{c}{\bf En-Ro} & \multicolumn{2}{c}{\bf En-Ru}  \\
\cmidrule(lr){2-3} \cmidrule(lr){4-5} 
& \bf BLEU &\bf TER &\bf BLEU &\bf TER \\
\midrule
\multicolumn{5}{c}{\it Existing Baselines} \\ 
XLM-R\scriptsize~\citep{conneau2019unsupervised} &35.6  &- &- &-\\
mRASP\scriptsize~\citep{lin2020pre} &37.6  &- &- &-\\
mBART\scriptsize~\citep{Liu:2020mbart} &37.7  &- &- &-\\ \midrule
\multicolumn{5}{c}{\it Our Implemented Systems} \\
Vanilla NMT &33.7 &48.6 & 28.8 &61.6  \\
~~+ PT &37.7 & 45.0 &31.6 &58.5 \\  \hdashline 
~~+ BT &38.4 & 45.0 &31.1 & 59.2 \\ 
~~+ BT + PT &41.2  & 42.6 &33.2  &57.1\\   \hdashline 
~~+ Tagged BT &38.6 & 44.9  &31.2 &59.3 \\ 
~~+ Tagged BT + PT &\bf 41.6 & \bf 42.1 &\bf 33.6 &\bf 56.5 \\   
\bottomrule
\end{tabular}}
\caption{Translation quality on the En-Ro and En-Ru benchmarks. ``+'' means incorporating PT and (Tagged) BT into NMT models. }
\label{tab:main}
\end{table}

The results are shown in Table~\ref{tab:main}. 
We use the vanilla model as our baselines, which are trained on original datasets with random initialization. 
Besides, we report results on existing PT models as our strong baselines, including XLM-R, mRASP, mBART.
As seen, PT can significantly improve the translation quality in all cases compared with vanilla baselines (averagely +2.5 BLEU), which is consistent with (or better than) existing PT models (37.7 vs. 35.6$\sim$37.7 BLEU). 
Furthermore, two BT methods\footnote{Tagged BT is to add a special token at the beginning of each back-translated source sentence.} (i.e. BT and Tagged BT) perform closely, which improves the standard NMT models by +3.5/+3.7 BLEU points on average. 
Simply combining them (+BT+PT) can further boost performances for NMT models across different sizes of datasets, showing the robustness and effectiveness of this approach.
Encouragingly, the combination of Tagged BT and PT performs better than the simple one, leading to state-of-the-art performances on the two benchmarks.
Similar tendencies are observed in terms of the TER scores.
The above results illustrate the better complementarity between PT and Tagged BT on improving translation quality for NMT models.

\begin{table}[t]
\centering
\begin{tabular}{lrrr}
\toprule
\bf Model & {\bf All} & {\bf ~~Src} & {\bf ~~Tgt} \\
\midrule
Vanilla &33.7 &29.4  &38.3 \\
~~+ PT &37.7 &33.8 &42.0 \\
\hdashline
~~+ BT &38.4 &31.5 &45.4 \\
~~+ BT + PT &41.2   &33.3 &48.6  \\ 
\hdashline 
~~+ Tagged BT & 38.6 &31.9 & 45.6 \\
~~+ Tagged BT + PT &\bf 41.6 &\bf 34.8  & \bf48.7 \\ 
\bottomrule
\end{tabular}
\caption{Translation quality of source-original and target-original sentences on the En-Ro benchmark. ``Src'' and ``Tgt'' respectively denote the sub-testsets of source-original and target-original while ``All'' means the whole testset.}
\label{tab:translationese}
\end{table}

\subsection{Analysis}
We conducted extensive analyses to better understand the improvement of our approach. All results are reported on the En-Ro benchmark.

\paragraph{Effects of Sentence Type}
Recent studies have shown that the evaluation of BT is sensitive to the sentences types, thus we report BLEU scores on the subsets of source-original (Src-Ori) and target-original (Tgt-Ori) datasets~\citep{zhang2019effect,liu-etal-2021-copying,wang-etal-2021-language-coverage}.\footnote{Src-Ori denotes the testing data originating in the source language, while Tgt-Ori denotes the data translating from the target language.}
Generally speaking, the translation of Src-Ori is more important than that of Tgt-Ori for practical NMT systems~\citep{graham-etal-2020-statistical}, thus its performance should be taken seriously.
As shown in Table~\ref{tab:translationese}, PT performs better on Src-Ori than BT (33.8 vs. 31.9 BLEU) while BT achieves higher scores on Tgt-Ori than PT (45.6 vs. 42.0 BLEU). 
Besides, simply combining PT and BT can improve the translation quality on both Src-Ori and Tgt-Ori sentences, but the improvement of Src-Ori is lower than only using PT.
By introducing tagged BT, the model can achieve better performance than the simple one, especially on source-original sentences. 
{\bf Takeaway:} {\em 1) PT and BT complementary in terms of originality of sentences; 2) Tagged BT can alleviate the bias of translating Tgt-Ori sentences which is significant to practical NMT systems.}

\begin{table}[t]
\centering
\begin{tabular}{lrrr}
\toprule
\bf Model & {\bf All} & {\bf Low} & {\bf High} \\
\midrule
Vanilla & 62.8 & 48.5 &64.6 \\
~~+ PT & 65.8 & 58.2 &66.7 \\
\hdashline
~~+ BT &65.9 &57.5 &67.1 \\
~~+ BT + PT &67.8 &60.8 & 68.8 \\ 
\hdashline 
~~+ Tagged BT & 66.1 &57.5 & 67.3 \\
~~+ Tagged BT + PT & \bf 68.3 &\bf 61.8 &\bf 69.1 \\ 
\bottomrule
\end{tabular}
\caption{F-measure of word translation according to frequency on the En-Ro benchmark. ``Low'' and ``High'' respectively denote the buckets of low- and high-frequency words while ``All'' means the whole words in the test set. Simply combining PT and BT improves the model performance, while adding tags to BT data further improves}
\label{tab:frequency}
\end{table}

\paragraph{Effects of Word Frequency}
Data augmentation is an effective way to improve the translation quality of low-frequency words~\citep{sennrich2016improving}. 
Thus, we compare the performance of the models on translating different frequencies of words. Specifically, we employed {\it compare-mt}~\citep{neubig-etal-2019-compare} to calculate the f-measure of translating low- and high-frequency words ($<$50 vs. $\geq$50). 
As shown in Table~\ref{tab:frequency}, PT improves more on translating low-frequency words (58.2 vs. 57.5 scores) while BT performs better on high-frequency words (67.3 vs. 66.7 scores). 
Furthermore, the combination of PT and tagged BT achieves the best performance on both low- and high-frequency words, leading to an overall improvement on the whole words.
Similar phenomenons can be observed by combining self-training and BT~\citep{ding-etal-2021-rejuvenating}. 
{\bf Takeaway:} {\em 1) PT and BT complementary in terms of frequency of words; 2) Tagged BT are more complementary to PT on lexical translation.}

\section{Conclusion and Future Works}
This paper broadens the understandings of pre-training (PT) and back-translation (BT).
We propose two probing tasks to investigate the impact of PT and BT on each NMT module and find that PT is more beneficial to the encoder while BT mainly improves the decoder.
Experimental results on the WMT16 English-Romanian and English-Russian benchmarks show that PT is nicely complementary to BT.
We also demonstrate that Tagged BT (i.e., adding tags to BT data) can further improve the complementarity of translating source-original sentences and low-frequency words.

In the future, we would like to apply curriculum learning~\citep{liu-etal-2020-norm,Zhan_Liu_Wong_Chao_2021,ding-etal-2021-progressive} to better organize the learning of PT and BT. It is also worthwhile to explore other kinds of methods utilizing monolingual data (e.g., self-training~\citep{zhang-zong-2016-exploiting,He2020Revisiting,jiao-etal-2021-self}) and validate the findings on practical NMT systems~\citep{transmart} and more generation tasks~\citep{liu2021understanding}.

\section*{Acknowledgement} 
This work was supported in part by the Science and Technology Development Fund, Macau SAR (Grant No. 0101/2019/A2), and the Multi-year Research Grant from the University of Macau (Grant No. MYRG2020-00054-FST). We thank the anonymous reviewers for their insightful comments.

\bibliography{emnlp2021}
\bibliographystyle{acl_natbib}
\end{document}